# Analysis and Optimization of fastText Linear Text Classifier.

Vladimir Zolotov and David Kung

IBM T. J. Watson Research Center, Yorktown Heights, NY, USA

zolotov@us.ibm.com, kung@us.ibm.com

## Abstract

The paper [1] shows that simple linear classifier can compete with complex deep learning algorithms in text classification applications. Combining bag of words (BoF) and linear classification techniques, fastText [1] attains same or only slightly lower accuracy than deep learning algorithms [2-9] that are orders of magnitude slower. We proved formally that fastText can be transformed into a simpler equivalent classifier, which unlike fastText does not have any hidden layer. We also proved that the necessary and sufficient dimensionality of the word vector embedding space is exactly the number of document classes. These results help constructing more optimal linear text classifiers with guaranteed maximum classification capabilities. The results are proven exactly by pure formal algebraic methods without attracting any empirical data.

## 1. Introduction

Text classification is a difficult important problem of Computational Linguistics and Natural Language Processing. Different types of neural networks (deep learning, convolutional, recurrent, LSTM, neural Turing machines, etc.) are used for text classification, often achieving significant success.

Recently, a team of researchers (A. Joulin, E. Grave P. Bojanowski, T. Mikolov) [1] experimentally has shown that comparable results can be achieved by a simple linear classifier. Their tool fastText [1] can be trained to the accuracy achieved with more complex deep learning algorithms [2-9], but orders of magnitude faster, even without using a high-performance GPU.

Exceptional performance of fastText is not a big surprise. It is a consequence of its very simple classification algorithm, and highly professional implementation in C++. High accuracy of very simple fastText algorithms is a clear indicator that the text classification problem is still not understood well enough to construct really efficient nonlinear classification models.

Because of very high complexity of nonlinear classification models, their direct analysis is too difficult. A good understanding of simple linear classification algorithms like fastText is a key to constructing good nonlinear text classifiers. That was the main motivation for analyzing the fastText classification algorithm. On the other hand, the simplicity of fastText makes very conducive for formal analysis. In spite of its simplicity, fastText combines several very important techniques: bag of words (BoW), representation of words as vectors of linear space, and linear classification. Therefore, a thorough formal analysis of fastText can further our understanding of other text classification algorithms employing similar basic techniques.

We obtained the following main results:

- The linear hidden layer of fastText model is not required for improving classification accuracy. We formally proved that any fastText type classifier can be transformed into an equivalent classifier without a hidden layer.
- The sufficient number of dimensions of the vector space representing document words is equal to the number of the document classes.
- Any fastText classifier recognizing *N* classes of documents can be algebraically transformed into an equivalent classifier with word vectors selected from *N*-dimensional linear space.

- In the general case, the minimum dimensionality of word vectors is the number of the document classes. It means, it is possible to construct a text classification problem with *N* classes of documents such that, for word vectors of *N-1* dimensional space, there is no fastText type classifier that correctly recognizes all classes. However, there exists fastText type classifier with *N*-dimensional word vectors, that can perform the required classification correctly.
- By simple modification of the classification algorithm, it is possible to reduce the necessary and sufficient dimensionality of the vector space of word representations by 1.

The above facts are proven using formal algebraic transformations. Therefore, these conclusions are exact and fully deterministic.

The proven theoretical facts have practical value. From them it follows that increasing length of word vectors beyond the number of document classes cannot improve the classification accuracy of linear BoW classifier. On other hand, if word vectors have fewer dimensions than the number of the document classes, we may fail to achieve the maximum possible accuracy. Besides, we see that by adding a hidden linear layer we cannot improve the accuracy of linear BoW classifier. According to the proven facts, an LBoW text classifier guaranteeing maximum achievable accuracy has well defined structure: word vectors with as many coordinates as the number of document classes to be recognized, and no hidden layer. It means that knowing the number of the document classes and the number of the dictionary words, the best LBoW classifier can be constructed automatically.

We show how any given linear BoW classifier is transformed into a simpler one without loss of the classification accuracy. The word vectors of the simpler equivalent classifier are computed from the word vectors and the parameters of the hidden layer of the precursor classifier.

Our analysis is based on the following idea. As soon as the classification is purely linear, and is performed by projecting a linear combination of word vectors onto *N*-dimensional space of the document classes, the word vectors can be directly selected from that *N*-dimensional space. This makes the projection trivial, and therefore, unnecessary.

For getting higher accuracy, fastText uses some other tricks: filtering rare words, considering letter n-grams (sub-words) [2], word n-grams, etc. These tricks improve the classification accuracy, but their effects are not critical for formal analysis of fastText classification model. Therefore, in our analysis we do not consider those tricks.

Our analysis is focused on the structure of classifiers, but does not consider explicitly how easy they can be trained, and how good is training convergence. However, from general point of view, the simpler the classifier, the fewer the number of parameters to be learned, the easier to train the classifier, and the less the risk of overfitting. Therefore, elimination of the hidden layer and reduction of dimensionality of word vectors should improve both the speed of recognition and training, and the convergence of training algorithms.

## 2. Background

### 2.1. Linear Bag of Words classifier

The fastText classifier presented in [1] is called here a Linear Bag Of Words (LBoW) classifier to emphasize the fact that it uses linear technique both for combining the word vectors into the vector representing the document, and for computing the classification criterion.

The structure of fastText LBoW classifier is shown in Fig.1. Document words $w_i$ are represented with *n*-dimensional word vectors $x_i$.

First, the classifier computes a document vector *y*. This vector *y* is a linear bag of words of the document. It is computed by averaging all document word vectors $x_i$:

$$y = \frac{1}{N}\sum_{i=1}^{N} x_i \quad (1)$$

Here, *N* is the number of words $w_i$ in the document. A document word $w_i$ is a word occurrence. Each word vector is taken in the averaging the same number of times as the word occurred in the document.

The document vector *y* is supplied to the input of the hidden layer, where it is multiplied by the matrix *B* of the hidden linear layer to get a classification vector *z*:

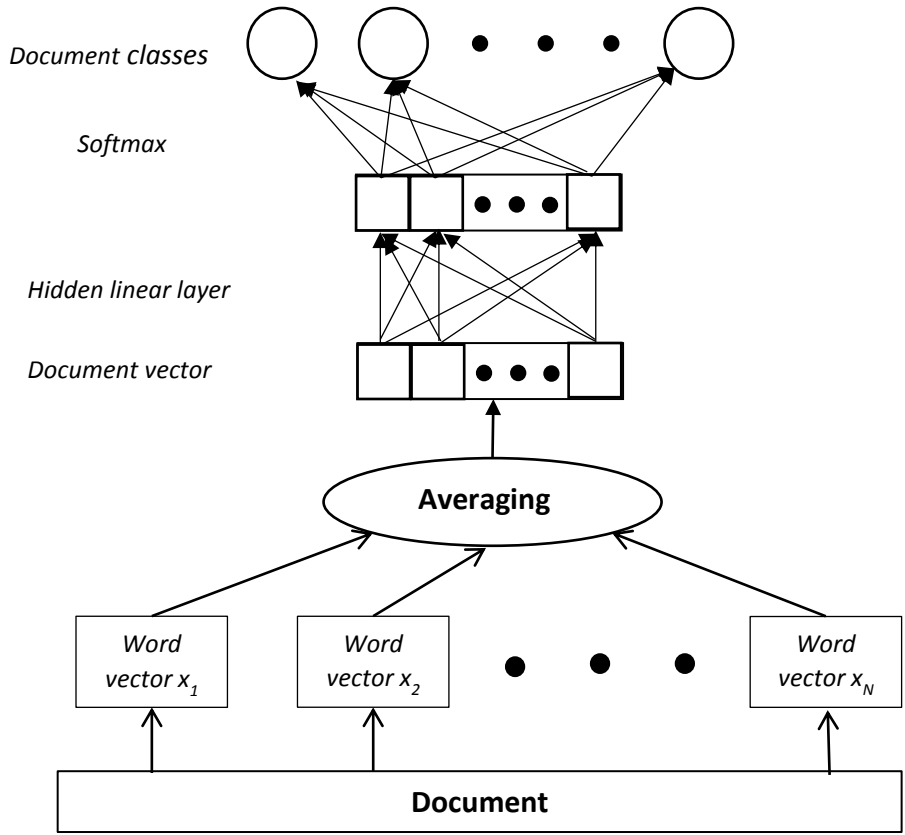

**Figure 1.** LBoW classifier with hidden linear layer

$$z = \begin{pmatrix} z_1 \\ \vdots \\ z_m \end{pmatrix} = \begin{bmatrix} b_{1,1} & \cdots & b_{1,n} \\ \vdots & \ddots & \vdots \\ b_{m,1} & \cdots & b_{m,n} \end{bmatrix} \begin{pmatrix} y_1 \\ \vdots \\ y_n \end{pmatrix} = B \cdot y \quad (2)$$

Here, *y* is an *n*-dimensional vector, *z* is an *m*-dimensional vector, *B* is an *m* by *n* matrix; *m* is the number of document classes (labels).

Classification is performed by computing the class probabilities with the *softmax* function:

$$p_j = \frac{e^{z_j}}{\sum_{k=1}^{m} e^{z_k}} \quad (3)$$

where $p_j$ is the predicted probability that the document belongs to the *j*-th class; $z_j$ and $z_k$, are the components of the classification vector *z*.

From the expression of the *softmax* function, it is obvious that predicted probabilities $p_i > p_j$, if and only if the corresponding classification vector coordinates $z_i > z_j$. Therefore, the classification can be performed by explicitly comparing the coordinates $z_k$ of the classification vector $z$ without computing the probabilities by the *softmax* function. The role of the softmax function is to make the classification function continuous and differentiable to be convenient for computing training gradient.

The described classification model computes the document vector *y* by averaging vectors of all word occurrences. However, our analysis is also valid for the case when the multiple word occurrences are ignored. The analysis is also valid if the averaging is replaced with any other linear combination of the document word vectors.

## 2.2. Classification problems and LBoW classifiers

For our analysis, we need the following formal definitions.

**Definition 1**. The LBoW classifier *C* with a hidden layer is a pair *(X,B)*, where *X* is a set of the word vectors $x_i$, and *B* is the matrix of the hidden layer.

**Definition 2.** Classification problem *P* is a quadruplet *(D, T, L, F)*, where *D* is a dictionary of words, *T* is a set of documents, *L* is a set of document classes (labels), and *F* is a function assigning classes (labels) to the documents.

**Definition 3.** We say, two classification problems $P_1 = (D_1, T_1, L_1, F_1)$ and $P_2 = (D_2, T_2, L_2, F_2)$ have the same dimensions (types) if both dictionaries $D_1$ and $D_2$ have the same number of words, and the both classification problems have the same number of classes, i.e. the sets $L_1$ and $L_2$ have the same number of elements.

In our analysis, we consider the classifiers that can be applied to the classification problems of the same dimensions. All those classifiers have the same number of word vectors representing the dictionary words. Also, their matrices $B_i$ of the hidden layers have the same number of rows, because all the problems of the same dimension have the same number of the document classes. The length of the vectors representing the words and the number of columns of the hidden layer matrix *B* can be different even for classifiers applicable to the same classification problem.

The accuracy of the LBoW classifier to which the classifier can be trained depends on dimensionality of its word vectors. Analysis of that dependence is the main goal of the following analysis.

## 2.3. Equivalency of classifiers

Our analysis uses a notion of equivalency of classifiers.

**Definition 4.** Two classifiers $C_1$ and $C_2$ are equivalent, if they can be applied to the same classification problems, and always assign the same classes to the same documents.

**Definition 5.** Two classifiers $C_1$ and $C_2$ strictly equivalent, if they can be applied to the same classification problems, and for the same document they always compute the same class probabilities.

It is obvious, that if two LBoW classifiers $C_1 = (X_1, B_1)$ and $C_2 = (X_2, B_2)$ are equivalent, then the sets $X_1$ and $X_2$ of the word vectors have the same number of elements, and the hidden layer matrices $B_1$ and $B_2$ have the same number of rows. However, in equivalent LBoW classifiers the vectors of word representations may be different, and may have different length. Also, the hidden layer matrices of equivalent LBoW classifiers may be different, and may have different number of columns.

## 3. Equivalent transformation of LBoW classifier

We transform a given LBoW classifier to a strictly equivalent one with a simpler structure. The following theorem defines this transformation and proves strict equivalency of the resulted classifier and its precursor.

**Theorem 1.** For any LBoW classifier with *m* document classes, there exists a strictly equivalent classifier with *m*-dimensional word vectors. That classifier has no hidden layer.

**Proof.** Combining formulas (1) and (2), we get the following formula for a classification vector computed by an LBoW classifier *C=(X,B)*:

$$z = B \cdot y = B \cdot \left(\frac{1}{N}\sum_{i=1}^{N} x_i\right) \quad (4)$$

where:

- *z* is the classification vector of a document *d*.
- $x_i$ are vectors representing words $w_i$ of the document *d*.
- *N* is the number of words in the document *d*.
- $y = \frac{1}{N}\sum_{i=1}^{N} x_i$ is the vector representing the document *d*.
- *B* is the matrix of the hidden layer of the classifier.

In this formula, the multiplication of the document vector *y* by the matrix *B* can be replaced with the multiplication of each word vector $x_i$ with the same matrix. Introducing vectors $\acute{x}_i = B \cdot x_i$ that formula can be rewritten as follows:

$$z = \frac{1}{N}\sum_{i=1}^{N} \acute{x}_i \quad (5)$$

This formula can be interpreted as an LBoW classifier *C'* without a hidden layer. The word vectors $x'_i$ of this classifier are obtained from the word vectors $x_i$ of the classifier *C=(X,B)* by multiplying them by the matrix *B*. The dimensionality of the word vectors $x'_i$ is exactly the number of the document classes.

In the constructed classifier, the classification vector *z* is the document vector: the average vector of the document word vectors $x'_i$. The classification is performed directly on the document vectors themselves.

Obviously, the constructed classifier *C'* computes the same classification vectors and therefore is strictly equivalent to its precursor. We denote LBoW classifiers without a hidden layer by *C(X)*, omitting the symbol for matrix *B* from the more general notation *C(X, B)*.

The structure of an LBoW classifier without a hidden layer is shown in Fig. 2.

## 4. Minimum dimensionality of word vectors of LBoW classifier

We proved that if a classification problem has *m* document classes, there exists an LBoW classifier with m-dimensional word vectors, and that classifier can achieve the accuracy of the best LBoW classifier. However, it still is not clear if the same accuracy can be achieved by a LBoW classifier with word vectors having fewer coordinates than the number of document classes. Now we show that it is not always possible. The minimum length of word vectors guaranteeing the maximum accuracy is equal to the number of the document classes.

First, we prove the following lemma.

**Lemma 1.** If two document vectors *y'* and *y* differ only by a positive scaling factor *a* (i.e. *y'=a\*y*), then any LBoW classifier assigns them the same class.

**Proof.** The classification vectors are computed by multiplying the document vectors with the matrix of the hidden layer. Therefore, the both classification vectors differ only with the same positive scaling factor *a*.

The *softmax* classifier computes the class with the index equal to the index of the maximum coordinate of the classification vector. Since, multiplying a vector by a positive scalar does not affect order relation between its coordinates, the both documents are classified the same way.

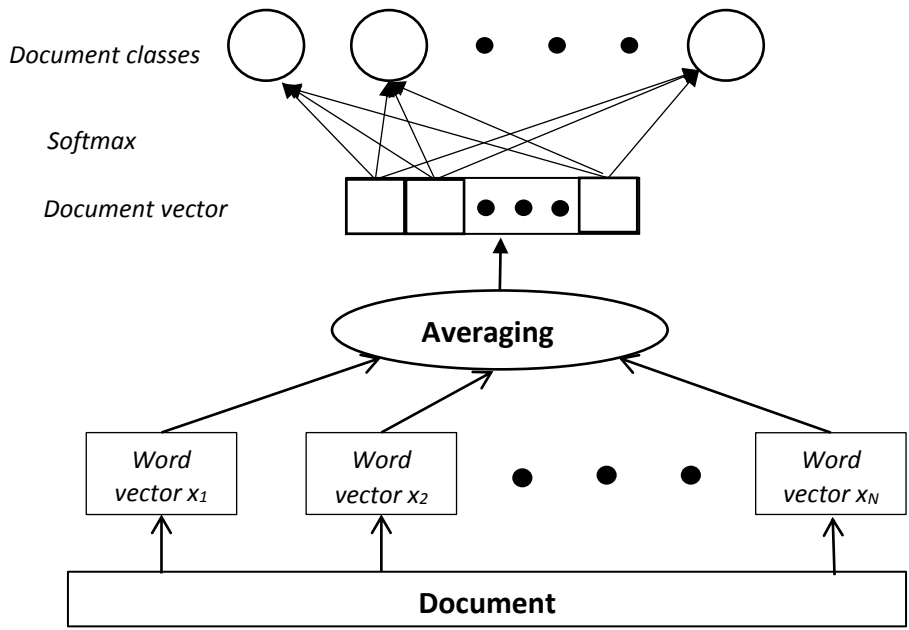

**Figure 2.** LBoW classifier without hidden layer

Now we are ready to prove the theorem on the minimum dimensionality of word vectors.

**Theorem 2.** There exists a classification problem with *m* document classes that can be solved by an LBoW classifier with *m*-dimensional word vectors, but cannot be solved with the same accuracy by any LBoW classifier with shorter word vectors.

**Proof.** First, consider a classification problem *P=(D, T, L, F)*, where:

- The dictionary *D* has exactly *m* words $\{w_1, w_2, .., w_m\}$.
- The set of documents *T* consists of all possible sequences of the dictionary words.
- The number of the document classes is exactly the number of the dictionary words.
- The function *F* assigns to a document *d* the label $l_i \in L$ such that the word $w_i$ is the most frequent word in the document.
- If several different words occur the same number of times in a document, and more often than any other word, then the document is assigned the class corresponding to the most frequent word with the lowest index.

Let an LBoW classifier $C_0$ represent each dictionary word $w_i$ with a vector $x_i$, with *i-th* coordinate equal 1, and all other coordinates 0. Since a document vector is an average of the document word vectors, the coordinates of the document vector are exactly the frequencies of the corresponding words in the document. The coordinate with the maximum value corresponds to the most frequent word. Comparing the coordinates of the document vector, the classifier always selects the correct document class. So, the constructed LBoW classifier solves the classification problem *P* exactly.

Now, consider an arbitrary classifier $C$ with the word vectors belonging to a linear $q$-dimensional vector space, where $q<m$. Since $m>q$, the $m$ word vectors of the classifier $C$ dictionary are not linearly independent. Therefore, there exist $m$ numbers $\{a_1, a_2, …, a_m\}$, satisfying the following equation:

$$\sum_{i=1}^{m} a_i x_i = 0 \quad (6)$$

where, $x_i$ are word vectors of the classifier, and some of the numbers $a_i$ are different from 0.

In this equation, at least two coefficients $a_i$ and $a_j$ are not 0.

Without loss of the generality, we can assume all coefficients $a_i$ are integers. By proper scaling, rational coefficients can be turned into integers. The case of irrational coefficients can be excluded, considering only word vectors with rational coordinates, which can approximate irrational number with any required accuracy. This is a reasonable assumption for any practical application.

First, consider the case when all coefficients $a_i$ are positive. Consider a document $d_0$ containing each word $w_i$ exactly $a_i$ times. By equation (6), the sum of the word vectors of this document is 0. Therefore, the document $d_0$ can be concatenated with any other document $d$ any number of times without changing the sum of the word vectors of the document $d$.

Let a document $d_1$ belong to a different class than the document $d_0$. By concatenating a sufficiently large number of copies of the document $d_0$ with the document $d_1$, we get the document $d_2$ where the most frequent word is the same as the one in the document $d_0$. Therefore, by the definition of the classification problem $P$, the document $d_2$ belongs to the same class as the document $d_0$, which is different from the class of the document $d_1$. However, both documents $d_1$ and $d_2$ have the same sum of their word vectors. Then, the vectors representing these documents differ only with a positive multiplier (the ratio of the length of the documents). According to Lemma 1, the classifier $C$ computes the same class for the both documents $d_1$ and $d_2$, which is incorrect. Therefore, this classifier is less accurate that the exact classifier $C_0$, constructed above.

The case of all negative coefficients $a_i$ does not require special consideration as by multiplying the equation (6) by -1, we obtain the equation where all non-zero coefficients are positive.

Now, consider the case when the non-zero coefficients in equation (6) have different signs. Moving the negative terms to the right and omitting all zero terms, we get the following:

$$\sum_{j=1}^{u} a_j x_j = \sum_{k=1}^{v} a_k x_k \quad (7),$$

where: all coefficients $a_j$ and $a_k$ are positive, and all word vectors $x_j$ and $x_k$ are different.

We construct two documents $d_l$ and $d_r$. The document $d_l$ has words $w_j$ corresponding to the word vectors $x_j$ from the left side of equation (7), and the document $d_r$ has the words $w_k$ corresponding to word vectors $x_k$ from the right side of equation (7). Each word $w_i$ occurs in its document exactly $a_i$ times. These documents belong to different classes because they have different words.

According to equation (7), the both documents have the same sums of the word vectors, and by Lemma 1, the classifier $C$ computes the same class for both documents $d_1$ and $d_2$, which is incorrect. Therefore, this classifier is less accurate that the exact classifier $C_0$, constructed above.

Thus, we proved that any classifier with word vectors having fewer coordinates than the number of the document classes, cannot solve the problem $P$ exactly while the classifier $C_0$ can do that. That proves the theorem.

## 5. Further reduction of dimensionality

By slight change of the classification algorithm, we further reduce dimensionality of word vectors of the LBoW classifiers with the structure defined in the Theorem 1. The idea is based on the following lemma.

**Lemma 2.** Let *C(X)* be an LBoW classifier without a hidden layer. Consider an LBoW classifier *C'=(X')*, with the same number of the word vectors. Let the word vectors $x'_i$ of the classifier *C'* be obtained from the word vectors $x_i$ of the classifier *C* by decrementing all coordinates of each vector $x_i$ by the same value $a_i$. (That value $a_i$ can be different for different vectors $x_i$.) Then the classifiers *C'* and *C* are strictly equivalent.

**Proof.** Let $A_i=(a_i, a_i, ..., a_i)$. By assumption, if the classifier *C* has a word vector $x_i$, then the corresponding word vector of the classifier *C'* is $x'_i = x_i - A_i$. Then the document vector *y'* of classifier *C'* is as follows:

$$\hat{y} = \frac{1}{N}\sum_{k=1}^{N} \acute{x}_k = \frac{1}{N}\sum_{k=1}^{N}(x_k - A_k) = \frac{1}{N}\sum_{k=1}^{N} x_k - \frac{1}{N}\sum_{k=1}^{N} A_k = y - A \quad (8)$$

where: *y* and *y'* are the vectors representing the same document in the classifiers *C* and *C'*; $x_k$ and $x'_k$ are the word vectors of the classifiers *C* and *C'* of that document; $A=(a, a, ..., a)$, and $a = \frac{1}{N}\sum_{k=1}^{N} a_k$.

Thus, the coordinates of the document vector *y'* of the classifier *C'* are computed by decrementing all coordinates of the document vector *y* of the classifier *C* by the same value *a*.

The *softmax* function does not change its value when all its arguments $x_j$ are decremented with the same value *a*:

$$\frac{e^{x_i-a}}{\sum_{j=1}^{m} e^{x_j-a}} = \frac{e^{x_i}}{\sum_{j=1}^{m} e^{x_j}} \quad (9)$$

Therefore, the classifier *C'* computes the same class probabilities as the classifier *C*, and those classifiers are strictly equivalent, which proves the lemma.

Now, for a classifier *C(X)* without a hidden layer, we construct a classifier *C'(X')* by decrementing its word vectors with their last coordinate. For each vector $x_i=(x_{i,1}, x_{i,2}, ..., x_{i,m-1}, x_{i,m})$ of the classifier *C(X)*, the classifier *C'(X')* has the word vector $x'_i=(x_{i,1}-x_{i,m}, x_{i,2}-x_{i,m}, ..., x_{i,m-1}-x_{i,m}, 0)$.

Since the last coordinate of all word vectors of the classifier *C'(X')* is 0, that 0 can be discarded without any loss of the information.

Of course, the new classifier *C'(X')* has to use a modified classification algorithm. The probabilities of all classes except the *m*-th one are computed by the following formula:

$$p_j = \frac{e^{y_j}}{1+\sum_{k=1}^{m-1} e^{y_k}} \quad (10)$$

The probability of the *m*-th class is computed by a different formula:

$$p_m = \frac{1}{1+\sum_{k=1}^{m-1} e^{y_k}} \quad (11)$$

Combining this fact with the Theorem 1, we get the following:

**Theorem 3**. For any LBoW classifier with *m* document classes, there exist an equivalent linear classifier with *m-1*-dimensional word vectors. This classifier has no hidden layer.

## 6. Conclusions

The proven theorems show that word vectors can hold all the information that the LBoW classifier can learn from the training documents, and that no additional layers are required.

There are text classification problems requiring word vectors with as many coordinates at the number of document classes.

Increasing the length of word vectors beyond the number of the document classes cannot improve the accuracy of a LBoW classifier.

The ways to improve accuracy are either to extend the dictionary by including word combinations or sub-words, or to introduce non-linear functions considering non-linearity of word interactions.

Any LBoW classifier similar to fastText can be transformed into an equivalent classifier without a hidden layer and with word vectors having as many coordinates as many document classes is required to recognize. The transformation is performed by explicit recalculation of the word vectors, and does not required any retraining or tuning the constructed classifier.